%% file: main.tex
\pdfoutput=1

\documentclass[11pt]{article}

\usepackage{acl}

\usepackage{times}
\usepackage{latexsym}

\usepackage[T1]{fontenc}

\usepackage[utf8]{inputenc}

\usepackage{microtype}

\usepackage{graphicx}
\usepackage{amsmath}
\usepackage{amssymb}
\usepackage{bbding}
\usepackage{subcaption, booktabs} 
\usepackage{color, colortbl}
\usepackage{soul}
\usepackage{enumitem}
\usepackage{makecell}
\usepackage{multirow}
\usepackage{array}
\usepackage{multicol}
\usepackage{url}
\usepackage{CJKutf8}
\usepackage{hyperref}
\usepackage[ruled]{algorithm2e}
\usepackage{hyperref}
\usepackage{diagbox}

\usepackage{xcolor}
\definecolor{dark-green}{RGB}{0, 128, 128}
\definecolor{dark-red}{RGB}{192, 0, 0}
\definecolor{dark-blue}{RGB}{59,82,212}

\newcommand{\data}{\textbf{\texttt{CIDER}}}

\title{Inconsistent dialogue responses and how to recover from them}

\author{First Author \\
  Affiliation / Address line 1 \\
  Affiliation / Address line 2 \\
  Affiliation / Address line 3 \\
  \texttt{email@domain} \\\And
  Second Author \\
  Affiliation / Address line 1 \\
  Affiliation / Address line 2 \\
  Affiliation / Address line 3 \\
  \texttt{email@domain} \\}

\newcommand{\vt}{$^{\blacklozenge}$}
\newcommand{\tencent}{$^{\heartsuit}$}

\author{Mian Zhang\vt\Thanks{~Work done as an intern at Tencent AI Lab.},
        Lifeng Jin\tencent,
        Linfeng Song\tencent,
        Haitao Mi\tencent \and
        Dong Yu\tencent \\
        \vt{}Virginia Tech \tencent{}Tencent AI Lab, USA \\
        mianz@vt.edu, lifengjin@global.tencent.com
}

\begin{document}
\maketitle
\begin{abstract}
One critical issue for chat systems is to stay consistent about preferences, opinions, beliefs and facts of itself, which has been shown a difficult problem. In this work, we study methods to assess and bolster utterance consistency of chat systems. A dataset is first developed for studying the inconsistencies, where inconsistent dialogue responses, explanations of the inconsistencies, and recovery utterances are authored by annotators. This covers the life span of inconsistencies, namely introduction, understanding, and resolution. Building on this, we introduce a set of tasks centered on dialogue consistency, specifically focused on its detection and resolution. Our experimental findings indicate that our dataset significantly helps the progress in identifying and resolving conversational inconsistencies, and current popular large language models like ChatGPT which are good at resolving inconsistencies however still struggle with detection.\footnote{The dataset and codebase are released at \href{https://github.com/mianzhang/CIDER}{https://github.com/mianzhang/CIDER}.}

\end{abstract}

\begin{CJK*}{UTF8}{gkai}

\input{intro.tex}

\section{Related work}
\input{related_work}

\section{Data collection}
\input{data_collection}

\section{Annotation guidelines}\label{sec:guideline}
\input{guidelines}

\section{Data overview}
\input{data_overview}

\input{experiment}

\input{conclusion}

\bibliography{consistency, paperbib}
\bibliographystyle{acl_natbib}

\appendix


\end{CJK*}
\end{document}

%% file: intro.tex
\section{Introduction}
For years, inconsistencies in human-to-chatbot conversations have been evident~\cite{Dziri2019-rr,qin-etal-2021-dont,ji2023survey}, even in the era of large language models~\cite{mundler2023self}. We categorize these inconsistencies as either extrinsic or intrinsic. \textit{Extrinsic} inconsistencies~\cite{rashkin2021increasing,santhanam2021rome} arise when there's a discrepancy between a statement and an external source of information, such as a knowledge base. On the other hand, \textit{intrinsic} inconsistencies~\cite{Dziri2019-rr,Nie2020-wk,zheng2022cdconv} occur within the dialogue itself. These can manifest in two ways: through an intra-utterance contradiction~\cite{zheng2022cdconv}, where a single sentence contains conflicting information, or a history contradiction~\cite{Nie2020-wk}, where a current statement conflicts with a previous one. Our study particularly addresses history contradictions, a persistent challenge in conversational models due to the nature of language modeling: models could forget what they said due to intervening context~\cite{roller2020recipes}. 


\begin{figure}[!t]
    \centering
    \includegraphics[width=1.0\columnwidth]{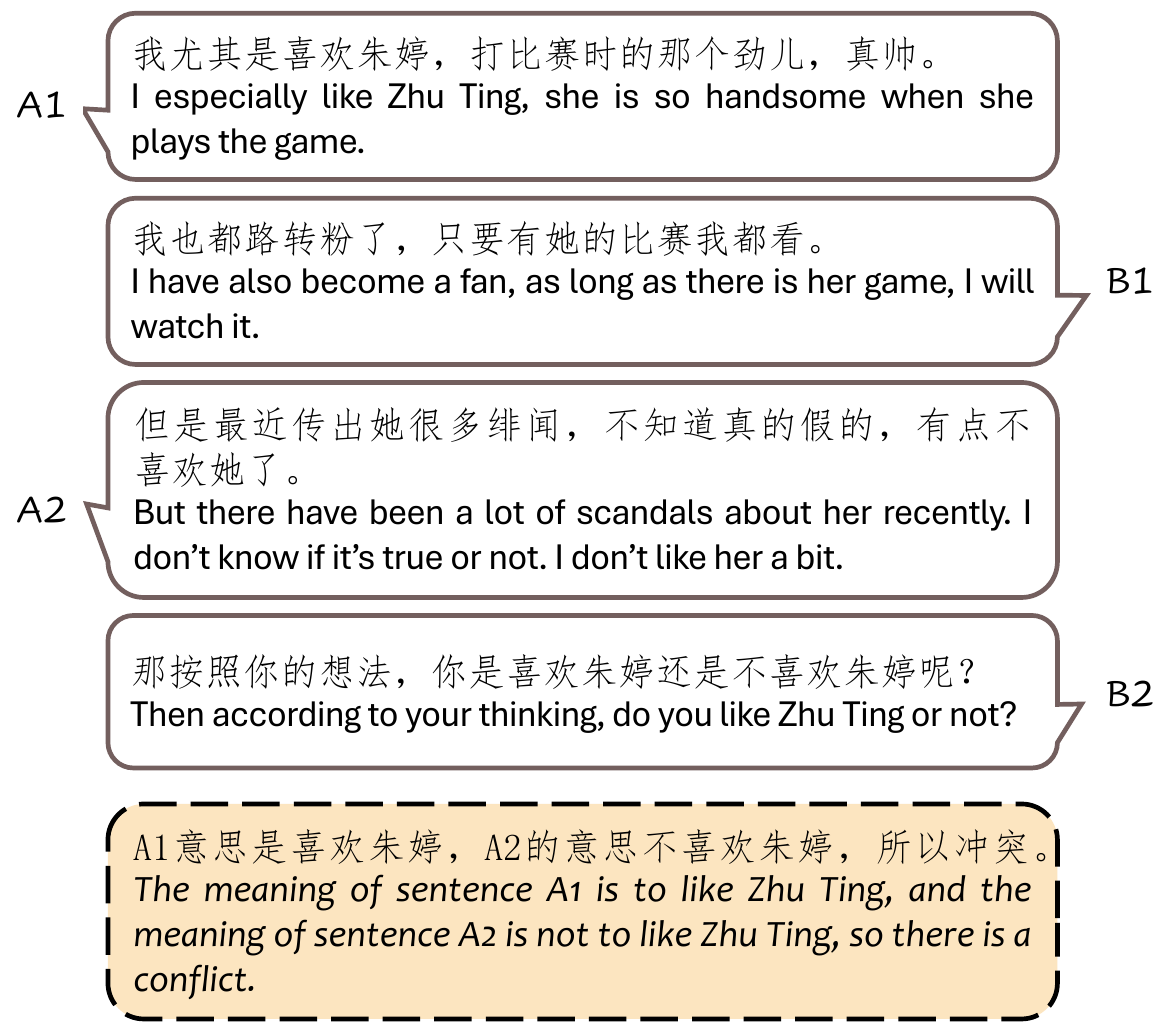}
    \caption{An instance in \data{} dataset. \{A, B\}$x$ denotes the $x$-th utterance of one of the two speakers (A or B). An inconsistent utterance (A2 in this case), an explanation of the inconsistency (the dotted box), and a clarification response (B2 in this case) are written for each dialogue.}
    \label{fig:intro}
\end{figure}

Researchers have been actively exploring how to resolve inconsistencies between utterances generated by conversational models in recent years. \citet{li2019don,rashkin2021increasing} has made progress in this domain by enhancing the training of these models, incorporating additional features and objectives to bolster self-consistency. Furthermore, \citet{lee2022factuality,su2022contrastive} introduced innovative decoding algorithms aimed at fostering greater coherence in utterances. These preemptive approaches are designed to mitigate conversational inconsistencies by reducing the likelihood of generating responses that contradict previous dialogue. However, these approaches cannot resolve the inconsistencies that do occur, possibly from the user or from model errors.
Therefore it's equally important to robustly address inconsistencies that do arise. Various remedial techniques have shown promise in other tasks, from grammar error correction~\cite{wu2023chatgpt} and moderating inappropriate dialogue content~\cite{zhang2023safeconv}, to generating clarifying questions in information retrieval~\cite{Zamani2020-oz} and conversational question answering~\cite{Guo2021-yp}. However, there seems to be a significant gap in the research when it comes to directly addressing inconsistencies that do arise between utterances.


In this work, we first propose a human-authored dataset with 27,180 dialogues to study the inconsistencies between utterances. At a high level, the dataset, called \textbf{\data{}}, covers the whole life span of in\textbf{\underline{\texttt{C}}}onsistencies, encompassing their \textbf{\underline{\texttt{I}}}ntroduction, un\textbf{\underline{\texttt{DE}}}rstanding, and \textbf{\underline{\texttt{R}}}esolution. Specifically, for each dialogue, annotators are first asked to write an utterance with inconsistent content regarding one utterance in the history to continue the conversation (A2 in Figure~\ref{fig:intro}), and then explain why the two utterances are inconsistent with natural language (the dotted box in Figure~\ref{fig:intro}), and finally provide a clarification response to continue the dialogue to resolve the inconsistency\footnote{The dialogues and annotation in the dataset are in Chinese. We also offer an English version translated by ChatGPT to facilitate research.} (B2 in Figure~\ref{fig:intro}). Given its large collection of inconsistent utterances paired with clarifying responses, \data{} stands out as a valuable resource for researching strategies to mitigate conversational inconsistencies. 


Utilizing the \data{} dataset, we conduct comprehensive experiments and analyses to study dialogue inconsistencies. Our findings underscore that \data{} can facilitate the development of robust inconsistency checkers compared to models trained on comparable public datasets. In addition, our research indicates that classic models like T5 and BART face challenges in adeptly resolving inconsistencies by providing clarifying responses. When assessing the proficiency of large language models (LLMs) in identifying and resolving conversational inconsistencies, we discerned two key points: 1) LLMs, when employed as inconsistency checkers, still leave much to be desired in terms of performance. 2) In contrast, as resolvers of inconsistency, LLMs exhibit a higher success rate compared to the fully supervised BART resolver.




 

%% file: related_work.tex
\paragraph{Consistency checking.}
Natural Language Inference (NLI)~\cite{hu2020ocnli, saha2020conjnli} is a task closely related to our work, where
a provided hypothesis is evaluated for its logical consistency with a given premise, with both presented in natural language. Within the context of dialogues, \citet{welleck2018dialogue} framed the consistency checking in dialogue as NLI and annotated binary consistency labels between dialogue-persona or persona-persona sentence pairs from the Persona-Chat dataset~\cite{zhang2018personalizing}. \citet{Dziri2019-rr} employed NLI models to assess topic coherence between a current response and the preceding dialogue history. Meanwhile, \citet{Shuster2022-is} delved into the issue of role confusion, where dialogue systems might inadvertently adopt the identity of the other party involved, and proposed a reranker trained with human judgments of identity consistency. The most relevant works are from \cite{Nie2020-wk} and \cite{zheng2022cdconv}, where they created datasets providing supervision for contradiction detection between conversational sentences. Our work distinguishes itself by providing more extensive annotations, including explanations and clarification responses.


\paragraph{Consistency resolving in dialogue.} 

To enhance the self-consistency of conversational models, \citet{rashkin2021increasing} employed controllable features, steering models towards generating more consistent responses. \citet{lee2022factuality} introduced factual-nucleus sampling and factuality-enhanced continued training to augment the reliability of language models during both decoding and training phases. \citet{Shuster2022-is} encouraged the conversational models to maintain an identity with the help of a role-playing accuracy classifier. \citet{li2019don} explored unlikelihood training~\cite{welleck2019neural} to curb inconsistencies in dialogue. However, given computational constraints, contemporary conversational models tend to rely predominantly on recent dialogue history when formulating responses. This predisposes them to produce content that may contradict earlier parts of the dialogue, especially distant sections. Generating clarification questions has emerged as a strategy to address communication breakdowns in dialogues, such as resolving ambiguities in a query during conversational information retrieval~\cite{Zamani2020-nq} or clarifying ambiguous user questions in conversational question answering~\cite{Guo2021-yp} scenarios. In this research, we propose an approach to recover from conversational inconsistencies by generating clarification questions, with the support of the proposed dataset.




\paragraph{Large language models.} 
Recent advancements in AI have been dominated by the rise of large language models, notably ChatGPT~\cite{ouyang2022training}, GPT-4~\cite{OpenAI2023GPT4TR} and others. They has shown that by scaling up language models, they can be equipped to tackle intricate tasks, such as question answering, machine translation, and numerical reasoning. In this study, leveraging the extensive annotations of our proposed dataset, \data{}, we assess these models' proficiency in detecting and addressing conversational inconsistencies.


%% file: data_collection.tex
The candidate conversations for annotation are sampled from two open-source conversation datasets: LCCC and NaturalConv.
LCCC~\cite{wang2020large} is a large collection of short conversations from the Chinese social media platform Weibo. NaturalConv~\cite{wang2021naturalconv} is an annotator-written dataset containing conversations around news items on topics like film and sports. They are different in content and style. LCCC conversations tend to be short in number of turns, and more in the style of daily chitchat. NaturalConv conversations, in contrast, are two to five times longer and contain more serious discussions about events in sports, films, and other areas. 
20,000 and 10,000 conversations are sampled from the LCCC and NaturalConv respectively for annotation. When sampling, conversations that are shorter than 4 turns or contain utterances shorter than 5 words are filtered out.

The sampled conversations are generally consistent, therefore the goal of data annotation is to create an alternative conversation that contains inconsistent utterances. To achieve this, we truncate the original conversation to create a common conversation context. For LCCC, the last utterance is truncated for inconsistent continuation writing; for NaturalConv, a random turn between 8 and $l - 4$%
\footnote{The last turns of NaturalConv tend to be goodbyes, therefore we choose to truncate before such utterances.}
and the following turns are chosen for truncation, where $l$ is the length of the conversation.

Finally, a specified source turn is sampled from the last turn or the turn before the last. This source turn is designated to be the source of the inconsistency where the following inconsistent continuation needs to form inconsistency with the utterance from the same speaker in this turn. 

\begin{table}[!t]
    \centering
    \resizebox{\columnwidth}{!}{
        \begin{tabular}{lllllll}
        \toprule
                                                      & \multicolumn{3}{c}{LCCC}       & \multicolumn{3}{c}{NaturalConv} \\
                                                      \cmidrule[0.5pt](rl){2-4}                       \cmidrule[0.5pt](rl){5-7}                
                                                      & Train          & Dev          & Test         & Train          & Dev          & Test                       \\
            \midrule
            \# of Convs           &    14,126     &  1,883  &  1,797    &  7,537   &  917     &   920    \\
             Ave. Cont. Len.  &  29.3    &  28.9  &  28.9   &   40.4     &   40.9     &   40.5    \\
             Ave. Exp. Len.     &  40.9   &  40.5  &  41.0  & 50.4   &  50.3  &   50.3  \\
             Ave. Res. Len.     &   16.2  &  16.1   &   16.1  &  20.3   &  20.1  &  20.0  \\
        \bottomrule
        \end{tabular}
    }
    \caption{Some basic statistics of the annotated datasets. Ave. Cont. Len. is the average continuation length in number of Chinese characters; Ave. Exp. Len. is the average explanation length; Ave. Res. Len. is the average resolution question length. They correspond to the outcome from the three annotation tasks.}
    \label{tab:data_overview}
\end{table}

\begin{table*}[!t]
    \centering
    \resizebox{0.9\textwidth}{!}{
        \begin{tabular}{lllllllllllll}
        \toprule
        & \multicolumn{6}{c}{\textit{Pair-Check}}       & \multicolumn{6}{c}{\textit{Diag-Check}}  \\
        \cmidrule[0.5pt](rl){2-7}                       \cmidrule[0.5pt](rl){8-13}           
        & \multicolumn{2}{c}{Train}          & \multicolumn{2}{c}{Valid}          & \multicolumn{2}{c}{Test}          & \multicolumn{2}{c}{Train}          & \multicolumn{2}{c}{Valid}          & \multicolumn{2}{c}{Test}       \\
        \cmidrule[0.5pt](rl){2-3} \cmidrule[0.5pt](rl){4-5} \cmidrule[0.5pt](rl){6-7} 
        \cmidrule[0.5pt](rl){8-9} \cmidrule[0.5pt](rl){10-11} \cmidrule[0.5pt](rl){12-13}  
        & \#Pos & \#Neg & \#Pos & \#Neg & \#Pos & \#Neg & \#Pos & \#Neg & \#Pos & \#Neg & \#Pos & \#Neg \\
        \midrule 
            STANCE                         & 1816        & 3959          & 195          & 446          & 346          & 644 
            & 1816  & 3959 & 195 & 446 & 346 & 644\\
            OCNLI                          & 14837       & 30601          & 1639          & 3409          & 900          & 2100    & 14837       & 30601          & 1639          & 3409          & 900          & 2100    \\
            CDConv                         & 2623       & 4373          & 880          & 1452          & 848          & 1484 
            & 2623  & 4373 & 880 & 1452 & 848 & 1484\\
            \data{}                         & 21663       & 53012          & 2800          & 6692          & 2717          & 6569  & 21663 & 21663 & 2800 & 2800 & 2717 & 2717\\
        \bottomrule
        \end{tabular}
    }
    \caption{Dataset statistics for checking tasks. Pos/Neg corresponds to label \textit{inconsistent}/\textit{consistent}.}
    \label{tab:data_check}
\end{table*}

%% file: guidelines.tex
The annotation process has been divided into three different tasks: inconsistent continuation, inconsistency explanation, and inconsistency resolution, which are required to be performed to each candidate conversation by one annotator when given a candidate conversation and a specified source turn.

\noindent \textbf{Inconsistent continuation.} The annotator first tries to create a natural continuation of the conversation by providing a possible utterance to the candidate conversation, but forms an inconsistency with the specified source utterance (A2 in Figure \ref{fig:intro} is the continuation, and A1 is the source.) The annotators are instructed to write the utterance with contradictory viewpoints, reasoning, and argumentation, instead of providing simple negation to the source utterance. For example, for the specified utterance \textit{I went to the supermarket yesterday.}, the continuation meeting the annotation requirement is \textit{I have been staying home for the past four days, not really wanting to go anywhere}, instead of \textit{I didn't go to the supermarket yesterday.}

\noindent \textbf{Inconsistency explanation.} After writing the continuation of the candidate conversation, the annotator is instructed to write down the rationale behind the created inconsistency (the dashed box in Figure \ref{fig:intro}). They are asked to follow this template when writing the rationale: \textit{The specified utterance means X, but the continuation utterance means Y, which is in contradiction with X.}, where the utterance meanings should be explicit. In the example above, the explanation one may write is \textit{The specified utterance indicates that I went out of my home yesterday, but the continuation utterance means that I didn't go out for many days including yesterday, which is in contradiction with the previous statement.}

\noindent \textbf{Inconsistency resolution.} Finally, the annotator provides another utterance to expose and question the inconsistency from a different party than the continuation party (B2 in Figure \ref{fig:intro}). The annotator is asked to write the resolution question naturally with the main purpose being clarifying the situation instead of complaining. They are also asked to try varying how the clarification question is raised, because the most intuitive way is asking by providing a binary choice. The resolution question for the example above is \textit{So were you home yesterday or did you go to the supermarket?}

Twelve examples collected from the two data sources and annotated by the authors were provided to the annotators along with the guidelines, which cover a number of common mistakes that the authors discovered in the trial annotation. The annotation project lasted two months, with six annotators%
\footnote{
The chosen provider created a qualification test based on the annotation guidelines for selecting annotators. The annotators with the highest agreement with the authors were then chosen as annotators. They then went through an online training session with the authors to align with the understanding of guidelines from the authors.
They were paid twice the local average monthly salary for their contributions.
} %
participating in the project from a commercial annotation provider, who was chosen amongst three providers based on the performance in the trial annotation task. The items for annotation were segmented into batches, each with 3000 conversations. The annotated items are checked first by quality assurance specialists from the annotation provider by batch, and then spot-checked by the authors with the acceptance rate setting at 95\%.%
\footnote{The spot-check rate is 10\%.}
Candidate conversations which are not possible to form inconsistencies, such as conversations containing mostly utterances of simple greeting or agreeing, are dropped in the annotation process.

%% file: data_overview.tex
After annotation, 17,806 conversations from LCCC and 9,374 conversations from NaturalConv have valid annotation. They are further split into train, dev and test sets, shown in Table \ref{tab:data_overview}. The average continuation and explanation lengths from LCCC conversations are substantially shorter than from NaturalConv, indicating the simple nature of social media conversations. The resolution question lengths are closer than the other lengths, showing that resolution questions tend to be less influenced by context and style.

%% file: experiment.tex

    
    


\begin{table*}[!t]
\begin{subtable}{\textwidth}
    \centering
    \resizebox{0.95\textwidth}{!}{
        \begin{tabular}{lllllllllllll}
        \toprule
                                            & \multicolumn{3}{c}{\textit{STANCE Test}}     & \multicolumn{3}{c}{\textit{OCNLI Test}} & \multicolumn{3}{c}{\textit{CDConv Test (Turn)}} & \multicolumn{3}{c}{\textit{\data{} Test (Turn)}}\\
                                            \cmidrule[0.5pt](rl){2-4}                      \cmidrule[0.5pt](rl){5-7}                 \cmidrule[0.5pt](rl){8-10}                 \cmidrule[0.5pt](rl){11-13}
                                            & Pre. & Rec. & F1 & Pre. & Rec. & F1 & Pre. & Rec. & F1 & Pre. & Rec. & F1 \\
            \midrule
            $\textrm{C}_{\textrm{STANCE}}^{Turn}$  & \textbf{72.8}          & \textbf{60.4}          & \underline{\textbf{66.0}}$_{\textcolor{dark-green}{\Uparrow14.3}}$          & 37.7  & 19.4          & \textit{25.7}           & 38.1 & 21.3          & \textit{27.4}          & 37.5          & 14.4          & \textit{20.8} \\
            $\textrm{C}_{\textrm{OCNLI}}^{Turn}$   & 31.6          & 36.1 & \textit{33.7}       & \textbf{72.9}          & 74.9          & \underline{\textbf{73.9}}$_{\textcolor{dark-green}{\Uparrow10.2}}$           & 51.3          & 37.3          & \textit{43.2}          & 35.7          & 37.4          & \textit{36.5}$_{\textcolor{dark-blue}{\uparrow1.4}}$ \\
            $\textrm{C}_{\textrm{CDConv}}^{Turn}$  & 41.8          & 8.1          & \textit{13.6}          & 40.9          & 15.0          & \textit{22.0}           & \textbf{56.3}          & \textbf{72.9}          & \underline{\textbf{63.5}}$_{\textcolor{dark-green}{\Uparrow14.7}}$          & 29.8          & 42.8          & \textit{35.1} \\
            \midrule
            $\textrm{C}_{\data{}}^{Turn}$     & 61.0 & 44.8          & \textit{51.7}$_{\textcolor{dark-blue}{\uparrow18.0}}$  & 30.7          & \textbf{76.2} & \textit{63.7}$_{\textcolor{dark-blue}{\uparrow38.0}}$  & 37.7          & 69.3 & \textit{48.8}$_{\textcolor{dark-blue}{\uparrow5.6}}$ & \textbf{76.2} & \textbf{69.3} & \underline{\textbf{72.6}}$_{\textcolor{dark-green}{\Uparrow36.1}}$ \\
        \bottomrule
        \end{tabular}
    }
    \caption{Performance of \textit{Pair-Check} checkers.}
    \label{tab:pair_check}
\end{subtable}

\vspace{0.2cm}

\begin{subtable}{\textwidth}
    \centering
    \resizebox{0.95\textwidth}{!}{
        \begin{tabular}{lllllllllllll}
        \toprule
                                            & \multicolumn{3}{c}{\textit{STANCE Test}}     & \multicolumn{3}{c}{\textit{OCNLI Test}} & \multicolumn{3}{c}{\textit{CDConv Test (Diag)}} & \multicolumn{3}{c}{\textit{\data{} Test (Diag)}}\\
                                            \cmidrule[0.5pt](rl){2-4}                      \cmidrule[0.5pt](rl){5-7}                 \cmidrule[0.5pt](rl){8-10}                 \cmidrule[0.5pt](rl){11-13}
                                            & Pre. & Rec. & F1 & Pre. & Rec. & F1 & Pre. & Rec. & F1 & Pre. & Rec. & F1 \\
            \midrule
            $\textrm{C}_{\textrm{STANCE}}^{Turn}$  & \textbf{72.8}  & \textbf{60.4}   & \underline{\textbf{66.0}}$_{\textcolor{dark-green}{\Uparrow20.4}}$   & 37.7  & 19.4     & \textit{25.7}  &   25.9   &  4.5  &  \textit{7.6}   &  48.4  & 21.8  & \textit{30.0} \\
            $\textrm{C}_{\textrm{OCNLI}}^{Turn}$  & 31.6      & 36.1    & \textit{33.7}       & \textbf{72.9}    & \textbf{74.9}   & \underline{\textbf{73.9}}$_{\textcolor{dark-green}{\Uparrow40.8}}$   &   46.6   &  37.6  &  \textit{41.6}$_{\textcolor{dark-blue}{\uparrow18.6}}$  &  52.5   &  42.7  &   \textit{47.1}$_{\textcolor{dark-blue}{\uparrow17.1}}$  \\
            $\textrm{C}_{\textrm{CDConv}}^{Diag}$  & 54.5      & 8.7  & \textit{15.0} & 31.5        & 16.2           & \textit{21.4}          & \textbf{62.5}          & \textbf{60.8}          & \textbf{61.7}$_{\textcolor{dark-green}{\Uparrow20.1}}$          & 61.3          & 8.3        & \textit{14.6} \\
            \midrule
            $\textrm{C}_{\data{}}^{Diag}$     & 38.8          & 55.2  &  \textit{45.6}$_{\textcolor{dark-blue}{\uparrow11.9}}$ &   33.7          & 32.4           & \textit{33.1}$_{\textcolor{dark-blue}{\uparrow7.4}}$          & 52.7          & 14.7          & \textit{23.0}          & \textbf{89.4}          & \textbf{91.6}        & \textbf{90.5}$_{\textcolor{dark-green}{\Uparrow43.4}}$ \\
        \bottomrule
        \end{tabular}
    }
    \caption{Performance of \textit{Diag-Check} checkers.}
    \label{tab:diag_check}

\end{subtable}
\caption{Performance of the checking tasks. The checker trained on dataset Y for task \textit{X-Check} is denoted as $\textrm{C}_{\textrm{Y}}^{X}$. The best result in each column is in bold. The best F1 score on each dataset is underscored and the points by which it exceeds the second best are shown by $\textcolor{dark-green}{\Uparrow}$. The transferring F1 scores on each dataset are in italics and the points by which they exceed the second best transferring score are shown by $\textcolor{dark-blue}{\uparrow}$. The performance of $\textrm{C}_{\textrm{STANCE}}^{Turn}$ and $\textrm{C}_{\textrm{OCNLI}}^{Turn}$ on \textit{STANCE Test} and \textit{OCNLI Test} in Table~\ref{tab:diag_check} is copied from Table~\ref{tab:pair_check}.}
\label{tab:check}
\end{table*}

\begin{table*}[!t]
    \centering
    \resizebox{0.95\textwidth}{!}{
        \begin{tabular}{lllllllllllll}
        \toprule
                                                      & \multicolumn{6}{c}{\textit{Merge}}            & \multicolumn{6}{c}{\textit{Pretrain}}   \\
                                                      \cmidrule[0.5pt](rl){2-7}                       \cmidrule[0.5pt](rl){8-13}
                                                      & \multicolumn{3}{c}{\textit{Pair-Check}}       & \multicolumn{3}{c}{\textit{Diag-Check}}   & \multicolumn{3}{c}{\textit{Pair-Check}}       & \multicolumn{3}{c}{\textit{Diag-Check}}  \\
                                                      \cmidrule[0.5pt](rl){2-4}                       \cmidrule[0.5pt](rl){5-7}                 \cmidrule[0.5pt](rl){8-10}                       \cmidrule[0.5pt](rl){11-13}          
                                                      & Pre.          & Rec.          & F1            & Pre.          & Rec.          & F1         
                                                      & Pre.          & Rec.          & F1            & Pre.          & Rec.          & F1            \\
            \midrule
            $\textrm{C}_{\textrm{\data{}}}$           & 76.2          & 69.3          & 72.6          & \textbf{89.4}          & 91.6          & 90.5        & 76.2          & 69.3          & 72.6          & 89.4          & 91.6          & 90.5         \\
            
            $ _{+ \textrm{CDConv}}$  &     \textbf{76.7}   &   72.5      & 74.6$_{\textcolor{dark-green}{\Uparrow2.0}}$          & \textbf{90.7}          & 91.9          & \textbf{91.3}$_{\textcolor{dark-green}{\Uparrow0.8}}$        &     76.4 &    \textbf{71.1}    & 73.7$_{\textcolor{dark-green}{\Uparrow1.1}}$          & 88.4          & 91.4     & 89.9$_{\textcolor{dark-red}{\Downarrow0.6}}$         \\
            
            $ _{+ \textrm{OCNLI}}$   & 70.1          & 77.4          & 73.6$_{\textcolor{dark-green}{\Uparrow1.0}}$          & 89.8          & 92.1          & 90.9$_{\textcolor{dark-green}{\Uparrow0.4}}$        &   \textbf{77.4}    &   70.7    &  \textbf{73.9}$_{\textcolor{dark-green}{\Uparrow1.3}}$      & 88.6          &     \textbf{93.1}      & \textbf{90.8}$_{\textcolor{dark-green}{\Uparrow0.3}}$         \\
            
            $ _{+ \textrm{STANCE}}$  & 72.4          & \textbf{77.9}          & \textbf{75.1}$_{\textcolor{dark-green}{\Uparrow2.5}}$          & 88.2          & \textbf{92.9}          & 90.5$_{\textcolor{dark-green}{\Uparrow0.0}}$        & 76.2          & 70.3          & 73.2$_{\textcolor{dark-green}{\Uparrow0.6}}$          &  87.3         &  92.7         & $89.9_{\textcolor{dark-red}{\Downarrow0.6}}$         \\
        \bottomrule
        \end{tabular}
    }
    \caption{Performance of checkers leveraging extra data on the test set of \data{}. The best are in bold. The relative increasing ($\textcolor{dark-green}{\Uparrow}$) and decreasing ($\textcolor{dark-red}{\Downarrow}$) points are calculated based on the performance of $\textrm{C}_{\textrm{\data{}}}$.}
    \label{tab:add_extra}
\end{table*}

\section{Consistency checking}\label{sec:check}
In this section, we experimentally verify whether the proposed \data{} could help the detection of inconsistency in conversation via two task settings: (1) checking the consistency between two sentences (\textit{Pair-Check}); (2) checking the consistency between an utterance and its preceding context (\textit{Diag-Check}). The (inconsistency) checker is initialized as  RoBERTa-base~\cite{liu2019roberta} with a linear binary classification head on the top. The input of the encoder for \textit{Pair-Check} is formatted as "\texttt{[CLS]} \{\textit{sentence 1}\} \texttt{[SEP]} \{\textit{sentence 2}\} \texttt{[SEP]}" while for \textit{Diag-Check}, "\texttt{[CLS]} \{\textit{context}\} \texttt{[SEP]} \{\textit{utterance}\} \texttt{[SEP]}", where the \texttt{[CLS]} and \texttt{[SEP]} are special tokens.

\paragraph{Baselines.} We compare \data{} with several related datasets:
\begin{itemize}
    \item CDConv~\cite{zheng2022cdconv}: a dataset with 12K dialogues for conversational contradiction detection. Compared to \data{}, CDConv covers another two types of contradiction: intra-sentence contradiction and role confusion. Each dialogue of CDConv contains two turns of utterances between a user and a bot and annotation of \textit{consistent} or \textit{inconsistent} between the replies of the bot.
    \item STANCE\footnote{\href{www.fudan-disc.com/sharedtask/AIDebater21/tracks.html}{www.fudan-disc.com/sharedtask/AIDebater21/tracks.html}}: a dataset for stance classification of articles of debating topics from online forums, where sentence pairs against each other are marked as \textit{inconsistent} and otherwise \textit{consistent}.
    \item OCNLI~\cite{hu2020ocnli}: a large-scale natural language inference (NLI) dataset, consisting of about 56,000 annotated sentence pairs. We regard sentence pairs with \textit{contradiction} label as \textit{inconsistent} and others as \textit{consistent}.
\end{itemize}

\paragraph{Implementation details.} For \data{}, when creating \textit{consistent} training instances of \textit{Pair-Check}, we regard all the utterances in the context of the same speaker without \textit{inconsistent} label as being consistent with the current response; and when creating the training instances of \textit{Diag-Check}, we drop current response with inconsistency and regard the previous response as being consistent with the context. Table~\ref{tab:data_check} shows the statistics of the datasets for these two checking tasks.

We adopt AdamW~\cite{loshchilov2017decoupled} to optimize models for 50 epochs with a learning rate of 1e-6 and a batch size of 16. We evaluate the model on the validation set at each epoch and keep the one with the best performance with an early stop patience of 3. All the results are averaged over three runs. Our experiments are run on two Nvidia V100 GPUs.

\paragraph{Results for \textit{Pair-Check}.} The performance of checkers trained on different datasets for \textit{Pair-Check} is demonstrated in Table~\ref{tab:pair_check}. For each checker, we show its performance on all the test sets of the evaluating datasets. 

There is a substantial distribution difference between the datasets with the checker trained on one dataset performing the best on the corresponding test set. $\textrm{C}_{\data{}}^{\textit{Turn}}$ has the largest exceeding F1 points over the second best, 36.1, indicating that the checker trained on other datasets is not good at detecting the consistency in the test set of \data{} and the training set of \data{} could provide useful supervision for it. Moreover, we compare the 0-shot transfer ability of checkers across the datasets. Results show that $\textrm{C}_{\data{}}^{\textit{Turn}}$ has the best transfer results on all the other three datasets, surpassing the second best by 18.0, 38.0, and 5.6 F1 points, respectively, demonstrating $\textrm{C}_{\data{}}^{\textit{Turn}}$ covering many similar linguistic phenomena in other datasets. On the whole, \textbf{\data{} provides robust supervision to check whether a pair of sentences are consistent, regardless of they are in a dialogue or not}.

\paragraph{Results for \textit{Diag-Check}.} The performance of the checkers trained on different datasets for \textit{Diag-Check} is demonstrated in Table~\ref{tab:diag_check}. The results of $\textrm{C}_{\textrm{CDConv}}^{Diag}$ and $\textrm{C}_{\data{}}^{Diag}$ indicates again the distribution difference between \data{} and CDConv also being significant for \textit{Diag-Check} task: \data{} do not cover role confusion and intra-sentence contradiction these two types of inconsistency while being much larger than CDConv. In addition, $\textrm{C}_{\data{}}^{Diag}$ outperforms $\textrm{C}_{\textrm{CDConv}}^{Diag}$ on \textit{STANCE Test} by 30.6 F1 points and on \textit{OCNLI Test} by 11.7 F1 points, which demonstrates better transferring ability of $\textrm{C}_{\data{}}^{Diag}$ to non-conversational scenarios. Therefore, along with the transferring results in Table~\ref{tab:pair_check}, \textbf{\data{} offers more transferable patterns for checking consistency, and may be complementary to CDConv in the conversational scenarios}. We also notice that $\textrm{C}_{\textrm{OCNLI}}^{Turn}$ is superior to $\textrm{C}_{\data{}}^{Diag}$ on \textit{CDConv Test (Diag)} and to $\textrm{C}_{\textrm{CDConv}}^{Diag}$ on \textit{\data{} Test (Diag)}, showing that the knowledge of inconsistency between sentences in OCNLI is also useful for the inconsistency checking in dialogue.  

\paragraph{Role of extra data.} We are interested in whether other datasets could improve the performance of $\textrm{C}_{\data{}}$. We leverage the training data of STANCE, OCNLI, and CDConv via two ways: 1) directly merging one of them into the training data of \data{} (\textit{Merge}); 2) pretraining the checker on one of them before training on \data{} (\textit{Pretrain}).

The results are presented in Table~\ref{tab:add_extra}. It's evident that \textbf{incorporating additional data generally enhances the overall performance of $\textrm{C}_{\data{}}$}, The only exception is that only pretraining on OCNLI could improve the checker for \textit{Diag-Check} task, which indicates better supervision signal from OCNLI for checking the inconsistency of an utterance. Compared with pretraining on extra data, directly merging them is superior, which could be ascribed to the phenomenon of catastrophic forgetting~\cite{kirkpatrick2017overcoming} of pretrained models. Moreover, \textit{Pair-Check} generally benefits from the extra datasets more than \textit{Diag-Check} because most of the extra datasets are intrinsically designed for checking of sentence pairs and in large quality so models could learn generalized patterns from them.

\begin{figure}[!h]
    \centering
    \includegraphics[width=0.9\columnwidth]{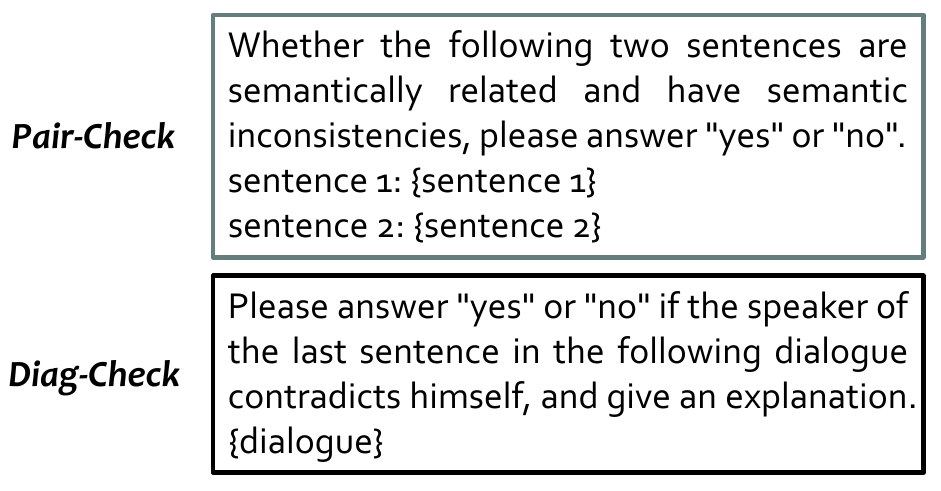}
    \caption{Prompts of checking tasks.}
    \label{fig:check_prompt}
\end{figure}

\paragraph{LLMs as consistency checker.}
We investigated the potential of large language models (LLMs) to function as robust consistency checkers. We pre-examine five human-crafted prompts for each task using a small-scale test set (50 instances) and select the best. The prompts applied for the checking tasks are illustrated in Figure~\ref{fig:check_prompt}. The evaluating LLMs are ChatGPT and GPT4\footnote{We use the versions gpt-3.5-turbo-0613 and gpt-4-0613 across our experiments.}. As shown in Table~\ref{tab:llm_check}, LLM-based checkers significantly lag behind the fully supervised $\textrm{C}_{\textrm{\data{}}}$, indicating that there is still much room for improvement. Moreover, the higher performance of GPT4 over ChatGPT underscores that larger LLMs possess a better capability to detect inconsistencies.

\begin{table}
    \centering
    \resizebox{\columnwidth}{!}{
        \begin{tabular}{lllllll}
        \toprule
                                                      & \multicolumn{3}{c}{\textit{Pair-Check}}       & \multicolumn{3}{c}{\textit{Diag-Check}} \\
                                                      \cmidrule[0.5pt](rl){2-4}                       \cmidrule[0.5pt](rl){5-7}                
                                                      & Pre.          & Rec.          & F1         & Pre.          & Rec.          & F1                       \\
            \midrule
            $\textrm{C}_{\textrm{\data{}}}$           & \textbf{76.2}          & 69.3          & \textbf{72.6}          & \textbf{89.4}          & \textbf{91.6}          & \textbf{90.5}         \\
             ChatGPT  &  42.0    & \textbf{79.0}    & 54.8  &  57.2        & 84.9         &  68.4      \\
             GPT4     &  49.9    & 76.2    & 60.3  & 68.8    & 82.1  & 74.8    \\
        \bottomrule
        \end{tabular}
    }
    \caption{Performance of LLMs on checking tasks.}
    \label{tab:llm_check}
\end{table}

\begin{table*}[!t]
    \centering
    \resizebox{0.85\textwidth}{!}{
        \begin{tabular}{llllllllll}
        \toprule
                                                       & & \multicolumn{4}{c}{\textit{Pair-Resolve}}       & \multicolumn{4}{c}{\textit{Diag-Resolve}}  \\
                                                      \cmidrule[0.5pt](rl){3-6}                       \cmidrule[0.5pt](rl){7-10}                       
                                    & \textbf{Model}   & BLEU      & R-1            & R-2          & R-L     & BLEU      & R-1            & R-2          & R-L       \\
            \midrule
            \#1 & $\textrm{T5}$             & 26.9          & 55.3          & 33.0        & 52.2       & 14.8          & 43.0          & 20.6          & 40.4           \\
            \#2 & $\textrm{BART}$          & 28.2$_{\textcolor{dark-green}{\Uparrow1.3}}$      & 57.2$_{\textcolor{dark-green}{\Uparrow1.9}}$   & 34.8$_{\textcolor{dark-green}{\Uparrow1.8}}$          & 53.7$_{\textcolor{dark-green}{\Uparrow1.5}}$      & 14.9$_{\textcolor{dark-green}{\Uparrow0.1}}$   & 43.7$_{\textcolor{dark-green}{\Uparrow0.7}}$          & 21.7$_{\textcolor{dark-green}{\Uparrow1.1}}$      & 41.0$_{\textcolor{dark-green}{\Uparrow0.6}}$           \\
            \midrule
            \#3 & $\textrm{T5}_{\textrm{oracle}}$   & 46.2      & 71.5        & 53.0          & 68.3          & 46.7          & 71.7          & 53.2          & 68.3  \\
            \#4 & $\textrm{BART}_{\textrm{oracle}}$ & 49.4$_{\textcolor{dark-green}{\Uparrow3.2}}$          & 74.4$_{\textcolor{dark-green}{\Uparrow2.9}}$        & 56.2$_{\textcolor{dark-green}{\Uparrow3.2}}$        & 70.7$_{\textcolor{dark-green}{\Uparrow2.4}}$         & 47.4$_{\textcolor{dark-green}{\Uparrow0.7}}$          & 72.4$_{\textcolor{dark-green}{\Uparrow0.7}}$  & 53.9$_{\textcolor{dark-green}{\Uparrow0.7}}$          & 68.7$_{\textcolor{dark-green}{\Uparrow0.4}}$           \\
            \midrule
            \#5 & ChatGPT & 14.3 & 45.2 & 22.2 & 41.4 & 5.3 & 29.8 & 9.9 & 26.9 \\
            \#6 & GPT4 & 10.8 & 42.7 & 20.2 & 38.0 & 4.1 & 28.0 & 9.8 & 24.2 \\
        \bottomrule
        \end{tabular}
    }
    \caption{Performance of resolvers on the test set of \data{}. The relative increasing ($\textcolor{dark-green}{\Uparrow}$) points of $\textrm{BART}$ ($\textrm{BART}_{\textrm{oracle}}$) are calculated based on the performance of $\textrm{T5}$ ($\textrm{T5}_{\textrm{oracle}}$).}
    \label{tab:resolve_baseline}
\end{table*}

\section{Consistency resolution}\label{sec:resolve}
Inconsistent responses of a conversational model could be detected by a consistency checker in advance, avoiding being exposed to users. However, inconsistent responses from a user can not be ignored by chat systems. The existence of inconsistent content may confuse the conversational model and induce undesired responses. Resolving the occurred inconsistency is necessary to maintain a smooth dialogue flow with clear semantics. The proposed \data{} dataset contributes to resolving the occurred inconsistency in a dialogue with \textit{clarification responses}, which is a valuable source to train an inconsistency resolution model.

We choose the base version of two representative conditional generative models to initialize the resolver: BART~\cite{lewis2019bart} and T5~\cite{raffel2020exploring}. They both follow an encoder-decoder structure and generate clarification responses in a sequence-to-sequence fashion: the conversational text with inconsistency is fed into the encoder and the clarification response is generated auto-aggressively by the decoder. Like the checking experiments in section~\ref{sec:check}, we consider two task settings: (1) generating a clarification response for a pair of inconsistent utterances (\textit{Pair-Resolve}); (2) generating a clarification response for a dialogue, of which the current response is inconsistent to the preceding context (\textit{Diag-Resolve}). The input of the encoder for \textit{Pair-Resolve} is formatted as "\texttt{[CLS]} \{\textit{utterance 1}\} \texttt{[SEP]} \{\textit{utterance 2}\} \texttt{[SEP]}" while for \textit{Diag-Resolve}, "\texttt{[CLS]} \{\textit{context}\} \texttt{[SEP]} \{\textit{response}\} \texttt{[SEP]}".

\paragraph{Implementation details.} We use the same optimization configuration of checkers to train the resolvers, except that a learning rate of 3e-4 is used for T5. BART and T5 are loaded with pretrained parameters from \citet{zhao2019uer} and \citet{shao2021cpt}, respectively. In decoding, we adopt Nucleus Sampling~\cite{holtzman2019curious} with top-0.90 probability mass across the experiments. 

\paragraph{Evaluation.} We use BLEU~\cite{papineni2002bleu} and ROUGE~\cite{lin2004rouge}, including ROUGE-1 (R-1), ROUGE-2 (R-2) and ROUGE-L (R-L), to measure the similarity between the generated text and the ground truth. 

\paragraph{Results.} According to rows \#1 and \#2 in Table~\ref{tab:resolve_baseline}, BART shows better performance in both \textit{Pair-Resolve} and \textit{Diag-Resolve} tasks than T5, indicating the pretrained parameters of BART are more suitable to inconsistency resolving. Meanwhile, the points of \textit{Pair-Resolve} are higher than those of \textit{Diag-Resolve}, which could be ascribed to \textit{Diag-Resolve} being a more difficult task than \textit{Pair-Resolve} because recognizing inconsistent contents between conversational context and a response is harder than between a pair of sentences. We also try appending \textit{explanations} to the input of the encoder to aid the generation process. Specifically, the input becomes "\texttt{[CLS]} \{\textit{utterance 1}\} \texttt{[SEP]} \{\textit{utterance 2}\} \texttt{[SEP]} \{\textit{explanation}\} \texttt{[SEP]}" for \textit{Pair-Resolve} and "\texttt{[CLS]} \{\textit{context}\} \texttt{[SEP]} \{\textit{response}\} \texttt{[SEP]} \{\textit{explanation}\} \texttt{[SEP]}" for \textit{Diag-Resolve}. The models with \textit{explanation} are denoted as $\textrm{T5}_{\textrm{oracle}}$ and $\textrm{BART}_{\textrm{oracle}}$, whose performances are shown at rows \#3 and \#4 in Table~\ref{tab:resolve_baseline}. We could see that $\textrm{T5}_{\textrm{oracle}}$ and $\textrm{BART}_{\textrm{oracle}}$ surpass T5 and BART by a significant margin, showing that with \textit{explanations} informing what inconsistency the input delivers, the models are able to produce clarification responses more semantically similar to the ground truth. Moreover, $\textrm{BART}_{\textrm{oracle}}$ performs better than $\textrm{T5}_{\textrm{oracle}}$ across all the metrics, demonstrating BART is better at exploiting \textit{explanations} to resolve semantic inconsistency.

\paragraph{Analysis.}\label{pa:ans} We go through 200 randomly selected instances (100 from \textit{Pair-Resolve} and 100 from \textit{Diag-Resolve}) of the best-performing BART resolver to 1) check whether the generated responses successfully clarify the inconsistent content and 2) explore the possible reasons that the clarification fails. The numbers of successful instances are presented in Table~\ref{tab:ans}. We could see \textbf{BART faces challenges in inconsistency resolution} and there is still large room for improvement. The higher success count for \textit{Pair-Resolve} compared to \textit{Diag-Resolve} indicates again that resolving inconsistencies between a response and its context poses greater challenges. We summarise the main types of failed clarification as follows: 

1. The resolver misses inconsistent content and just picks irrelevant semantic units to form a clarifying response. For instance, the user first says \textit{I want to buy a cup of coffee because I'm so sleepy.} and then \textit{Great, let's try Chinese tea!}. The resolver responds with \textit{Are you on earth sleepy or not?} This error type is common in \textit{Diag-Resolve} because long context contains irrelevant information that interferes with locating inconsistent content. 

2. The resolver includes the inconsistent content in the response but fails to form a fluent, contextual coherent response. For example, the user first says \textit{Are you free? I want you to do me a favor.} and then \textit{I am busy now.} and the resolver replies with \textit{Can you do a favor at all?}. In this case, the resolver misunderstands who is the subject of the action, thus providing a response incoherent to the context.

\begin{table}
    \centering
    \resizebox{0.75\columnwidth}{!}{
        \begin{tabular}{lcc}            
        \toprule
         & \multicolumn{2}{c}{\textbf{\#Succ. / \#Total}} \\
         \textbf{Model} & \textit{Pair-Resolve} & \textit{Diag-Resolve} \\
        \midrule
        $\textrm{BART}$ & 56 / 100 & 36 / 100 \\
        $\textrm{BART}_{\textrm{oracle}}$ & 91 / 100 & 82 / 100 \\
        \midrule
        ChatGPT & 76 / 100 &  64 / 100 \\
        GPT4 & 92 / 100 & 79 / 100 \\
        \bottomrule
        \end{tabular}
    }
    \caption{The number of successfully resolved instances.}
    \label{tab:ans}
\end{table}

\begin{figure}[!h]
    \centering
    \includegraphics[width=0.9\columnwidth]{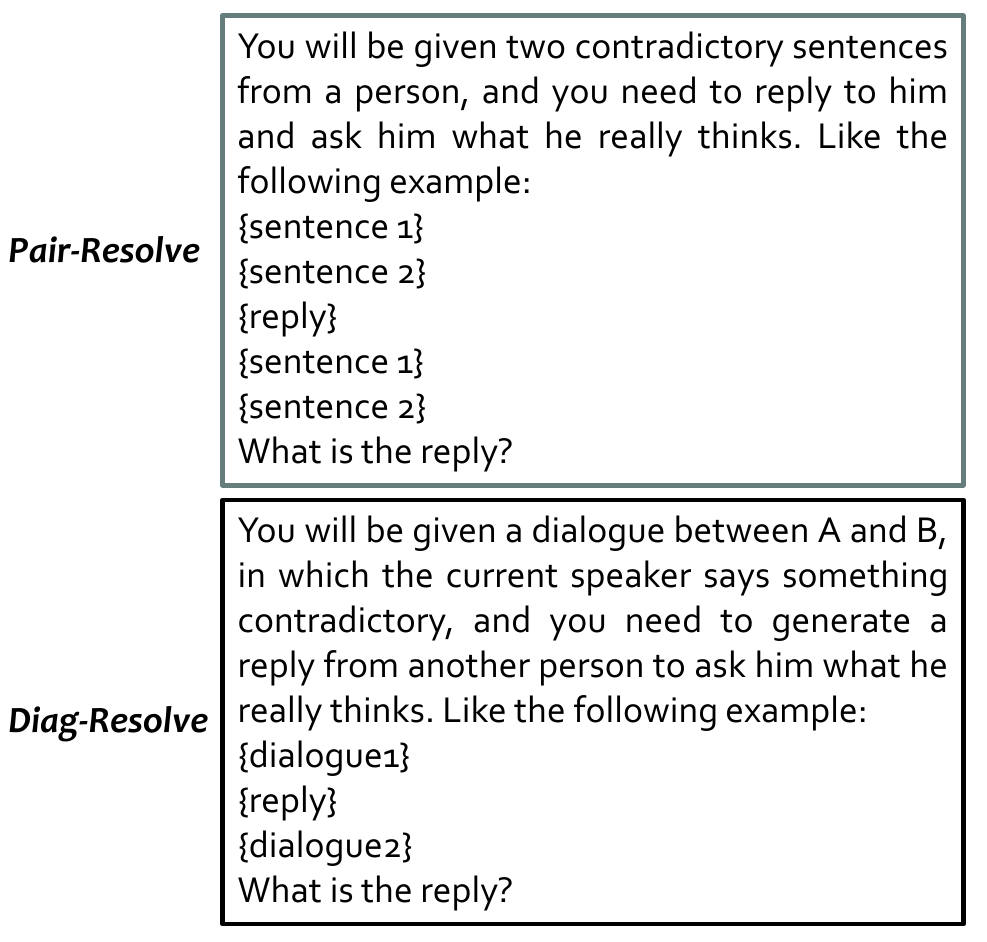}
    \caption{Prompts of resolving tasks.}
    \label{fig:resolve_prompt}
\end{figure}

\paragraph{LLMs as consistency resolver.}
We examine the consistency resolution ability of LLMs by asking LLMs to form a clarification response for the two resolving tasks via the prompts shown in Figure~\ref{fig:resolve_prompt} (one in-context example is included in the prompts to ensure a fixed output format).

We report automatic evaluation results in rows \#5 and \#6 of Table~\ref{tab:resolve_baseline}. On the selected instances in subsection \textbf{Analysis}, we conduct the same human evaluation of the generated clarification response of the LLMs and show the results in Table~\ref{tab:ans}. Results indicate that: \textbf{while ChatGPT and GPT4, both cutting-edge LLMs, score lower in BLEU and ROUGE compared to T5 and BART, they excel in addressing inconsistencies in dialogue history}, whose performance rivals that of the oracle resolvers. The lower BLEU and ROUGE scores of LLMs can be attributed to their tendency to produce more varied and extensive sentences. To illustrate, consider the reference clarification sentence: \textit{Do you really want to eat hot pot or barbecue?}. BART's response is, \textit{Do you really want to eat hot pot or not?}, whereas GPT4 offers, \textit{So, are you more attracted to hot pot, or does barbecue appeal to you more?}.

%% file: conclusion.tex
\section{Conclusion}
We present \data{}, a comprehensive dialogue dataset comprising 27,180 annotated dialogues to investigate conversational inconsistencies. The annotations of \data{} cover the whole life span of inconsistencies: the human-authored utterances with inconsistent content demonstrate the introduction of inconsistencies; the explanations help understand the inconsistencies; and the clarification responses exemplify how to resolve the inconsistencies. Through rigorous experiments and analysis, we show that \data{} significantly advance the detection and resolution of conversational inconsistencies, and large language models, ChatGPT and GPT4, exhibit commendable performance in resolving these conversational inconsistencies but struggle with identifying them.

\section*{Limitation}
Our work has following limitations:
\begin{itemize}

    \item Our proposed dataset emphasizes contradictions between utterances. For a truly effective system that detects or resolves inconsistencies, it is essential to incorporate resources that address other types of inconsistencies, such as intra-utterance or extrinsic discrepancies.

    \item We've currently evaluated the ability of LLMs to function as independent resolvers under specific prompts to generate clarification questions. The potential for these models to autonomously identify and clarify inconsistencies remains an intriguing avenue for future exploration. Moreover, while our evaluation of LLMs relies on the optimal prompts chosen from several human-crafted options, a more rigorous approach to prompt engineering could potentially yield superior outcomes.
\end{itemize}

\section*{Ethical consideration}
Our dataset, along with the LCCC~\cite{wang2020large} and NaturalConv~\cite{wang2021naturalconv} sources, have been cleaned to ensure no breaches of privacy (further details are available in their respective papers). All annotation guidelines (as detailed in Section~\ref{sec:guideline}) have received approval from the ethics review committee. We are confident that \data{} will play a pivotal role in crafting more human-friendly conversational models.

%% file: main.bbl
\begin{thebibliography}{37}
\expandafter\ifx\csname natexlab\endcsname\relax\def\natexlab#1{#1}\fi

\bibitem[{Dziri et~al.(2019)Dziri, Kamalloo, Mathewson, and
  Zaiane}]{Dziri2019-rr}
Nouha Dziri, Ehsan Kamalloo, Kory Mathewson, and Osmar Zaiane. 2019.
\newblock \href {https://doi.org/10.18653/v1/N19-1381} {Evaluating coherence in
  dialogue systems using entailment}.
\newblock In \emph{Proceedings of the 2019 Conference of the North {A}merican
  Chapter of the Association for Computational Linguistics: Human Language
  Technologies, Volume 1 (Long and Short Papers)}, pages 3806--3812,
  Minneapolis, Minnesota. Association for Computational Linguistics.

\bibitem[{Guo et~al.(2021)Guo, Zhang, Reddy, and Alikhani}]{Guo2021-yp}
Meiqi Guo, Mingda Zhang, Siva Reddy, and Malihe Alikhani. 2021.
\newblock {Abg-CoQA}: Clarifying ambiguity in conversational question
  answering.

\bibitem[{Holtzman et~al.(2020)Holtzman, Buys, Du, Forbes, and
  Choi}]{holtzman2019curious}
Ari Holtzman, Jan Buys, Li~Du, Maxwell Forbes, and Yejin Choi. 2020.
\newblock \href {https://openreview.net/forum?id=rygGQyrFvH} {The curious case
  of neural text degeneration}.
\newblock In \emph{8th International Conference on Learning Representations,
  {ICLR} 2020, Addis Ababa, Ethiopia, April 26-30, 2020}. OpenReview.net.

\bibitem[{Hu et~al.(2020)Hu, Richardson, Xu, Li, K{\"u}bler, and
  Moss}]{hu2020ocnli}
Hai Hu, Kyle Richardson, Liang Xu, Lu~Li, Sandra K{\"u}bler, and Lawrence Moss.
  2020.
\newblock \href {https://doi.org/10.18653/v1/2020.findings-emnlp.314} {{OCNLI}:
  {O}riginal {C}hinese {N}atural {L}anguage {I}nference}.
\newblock In \emph{Findings of the Association for Computational Linguistics:
  EMNLP 2020}, pages 3512--3526, Online. Association for Computational
  Linguistics.

\bibitem[{Ji et~al.(2023)Ji, Lee, Frieske, Yu, Su, Xu, Ishii, Bang, Madotto,
  and Fung}]{ji2023survey}
Ziwei Ji, Nayeon Lee, Rita Frieske, Tiezheng Yu, Dan Su, Yan Xu, Etsuko Ishii,
  Ye~Jin Bang, Andrea Madotto, and Pascale Fung. 2023.
\newblock Survey of hallucination in natural language generation.
\newblock \emph{ACM Computing Surveys}, 55(12):1--38.

\bibitem[{Kirkpatrick et~al.(2017)Kirkpatrick, Pascanu, Rabinowitz, Veness,
  Desjardins, Rusu, Milan, Quan, Ramalho, Grabska-Barwinska
  et~al.}]{kirkpatrick2017overcoming}
James Kirkpatrick, Razvan Pascanu, Neil Rabinowitz, Joel Veness, Guillaume
  Desjardins, Andrei~A Rusu, Kieran Milan, John Quan, Tiago Ramalho, Agnieszka
  Grabska-Barwinska, et~al. 2017.
\newblock Overcoming catastrophic forgetting in neural networks.
\newblock \emph{Proceedings of the national academy of sciences},
  114(13):3521--3526.

\bibitem[{Lee et~al.(2022)Lee, Ping, Xu, Patwary, Fung, Shoeybi, and
  Catanzaro}]{lee2022factuality}
Nayeon Lee, Wei Ping, Peng Xu, Mostofa Patwary, Pascale~N Fung, Mohammad
  Shoeybi, and Bryan Catanzaro. 2022.
\newblock Factuality enhanced language models for open-ended text generation.
\newblock \emph{Advances in Neural Information Processing Systems},
  35:34586--34599.

\bibitem[{Lewis et~al.(2020)Lewis, Liu, Goyal, Ghazvininejad, Mohamed, Levy,
  Stoyanov, and Zettlemoyer}]{lewis2019bart}
Mike Lewis, Yinhan Liu, Naman Goyal, Marjan Ghazvininejad, Abdelrahman Mohamed,
  Omer Levy, Veselin Stoyanov, and Luke Zettlemoyer. 2020.
\newblock \href {https://doi.org/10.18653/v1/2020.acl-main.703} {{BART}:
  Denoising sequence-to-sequence pre-training for natural language generation,
  translation, and comprehension}.
\newblock In \emph{Proceedings of the 58th Annual Meeting of the Association
  for Computational Linguistics}, pages 7871--7880, Online. Association for
  Computational Linguistics.

\bibitem[{Li et~al.(2020)Li, Roller, Kulikov, Welleck, Boureau, Cho, and
  Weston}]{li2019don}
Margaret Li, Stephen Roller, Ilia Kulikov, Sean Welleck, Y-Lan Boureau,
  Kyunghyun Cho, and Jason Weston. 2020.
\newblock \href {https://doi.org/10.18653/v1/2020.acl-main.428} {Don{'}t say
  that! making inconsistent dialogue unlikely with unlikelihood training}.
\newblock In \emph{Proceedings of the 58th Annual Meeting of the Association
  for Computational Linguistics}, pages 4715--4728, Online. Association for
  Computational Linguistics.

\bibitem[{Lin(2004)}]{lin2004rouge}
Chin-Yew Lin. 2004.
\newblock \href {https://aclanthology.org/W04-1013} {{ROUGE}: A package for
  automatic evaluation of summaries}.
\newblock In \emph{Text Summarization Branches Out}, pages 74--81, Barcelona,
  Spain. Association for Computational Linguistics.

\bibitem[{Liu et~al.(2019)Liu, Ott, Goyal, Du, Joshi, Chen, Levy, Lewis,
  Zettlemoyer, and Stoyanov}]{liu2019roberta}
Yinhan Liu, Myle Ott, Naman Goyal, Jingfei Du, Mandar Joshi, Danqi Chen, Omer
  Levy, Mike Lewis, Luke Zettlemoyer, and Veselin Stoyanov. 2019.
\newblock \href {https://arxiv.org/abs/1907.11692} {Roberta: A robustly
  optimized bert pretraining approach}.
\newblock \emph{ArXiv preprint}, abs/1907.11692.

\bibitem[{Loshchilov and Hutter(2019)}]{loshchilov2017decoupled}
Ilya Loshchilov and Frank Hutter. 2019.
\newblock \href {https://openreview.net/forum?id=Bkg6RiCqY7} {Decoupled weight
  decay regularization}.
\newblock In \emph{7th International Conference on Learning Representations,
  {ICLR} 2019, New Orleans, LA, USA, May 6-9, 2019}. OpenReview.net.

\bibitem[{M{\"u}ndler et~al.(2023)M{\"u}ndler, He, Jenko, and
  Vechev}]{mundler2023self}
Niels M{\"u}ndler, Jingxuan He, Slobodan Jenko, and Martin Vechev. 2023.
\newblock \href {https://arxiv.org/abs/2305.15852} {Self-contradictory
  hallucinations of large language models: Evaluation, detection and
  mitigation}.
\newblock \emph{ArXiv preprint}, abs/2305.15852.

\bibitem[{Nie et~al.(2021)Nie, Williamson, Bansal, Kiela, and
  Weston}]{Nie2020-wk}
Yixin Nie, Mary Williamson, Mohit Bansal, Douwe Kiela, and Jason Weston. 2021.
\newblock \href {https://doi.org/10.18653/v1/2021.acl-long.134} {{I} like fish,
  especially dolphins: Addressing contradictions in dialogue modeling}.
\newblock In \emph{Proceedings of the 59th Annual Meeting of the Association
  for Computational Linguistics and the 11th International Joint Conference on
  Natural Language Processing (Volume 1: Long Papers)}, pages 1699--1713,
  Online. Association for Computational Linguistics.

\bibitem[{OpenAI(2023)}]{OpenAI2023GPT4TR}
OpenAI. 2023.
\newblock \href {https://arxiv.org/abs/2303.08774} {Gpt-4 technical report}.
\newblock \emph{ArXiv preprint}, abs/2303.08774.

\bibitem[{Ouyang et~al.(2022)Ouyang, Wu, Jiang, Almeida, Wainwright, Mishkin,
  Zhang, Agarwal, Slama, Ray et~al.}]{ouyang2022training}
Long Ouyang, Jeffrey Wu, Xu~Jiang, Diogo Almeida, Carroll Wainwright, Pamela
  Mishkin, Chong Zhang, Sandhini Agarwal, Katarina Slama, Alex Ray, et~al.
  2022.
\newblock Training language models to follow instructions with human feedback.
\newblock \emph{Advances in Neural Information Processing Systems},
  35:27730--27744.

\bibitem[{Papineni et~al.(2002)Papineni, Roukos, Ward, and
  Zhu}]{papineni2002bleu}
Kishore Papineni, Salim Roukos, Todd Ward, and Wei-Jing Zhu. 2002.
\newblock \href {https://doi.org/10.3115/1073083.1073135} {{B}leu: a method for
  automatic evaluation of machine translation}.
\newblock In \emph{Proceedings of the 40th Annual Meeting of the Association
  for Computational Linguistics}, pages 311--318, Philadelphia, Pennsylvania,
  USA. Association for Computational Linguistics.

\bibitem[{Qin et~al.(2021)Qin, Xie, Huang, Chen, Xu, and
  Che}]{qin-etal-2021-dont}
Libo Qin, Tianbao Xie, Shijue Huang, Qiguang Chen, Xiao Xu, and Wanxiang Che.
  2021.
\newblock \href {https://doi.org/10.18653/v1/2021.emnlp-main.182} {Don{'}t be
  contradicted with anything! {CI}-{T}o{D}: Towards benchmarking consistency
  for task-oriented dialogue system}.
\newblock In \emph{Proceedings of the 2021 Conference on Empirical Methods in
  Natural Language Processing}, pages 2357--2367, Online and Punta Cana,
  Dominican Republic. Association for Computational Linguistics.

\bibitem[{Raffel et~al.(2020)Raffel, Shazeer, Roberts, Lee, Narang, Matena,
  Zhou, Li, and Liu}]{raffel2020exploring}
Colin Raffel, Noam Shazeer, Adam Roberts, Katherine Lee, Sharan Narang, Michael
  Matena, Yanqi Zhou, Wei Li, and Peter~J. Liu. 2020.
\newblock \href {http://jmlr.org/papers/v21/20-074.html} {Exploring the limits
  of transfer learning with a unified text-to-text transformer}.
\newblock \emph{J. Mach. Learn. Res.}, 21:140:1--140:67.

\bibitem[{Rashkin et~al.(2021)Rashkin, Reitter, Tomar, and
  Das}]{rashkin2021increasing}
Hannah Rashkin, David Reitter, Gaurav~Singh Tomar, and Dipanjan Das. 2021.
\newblock \href {https://doi.org/10.18653/v1/2021.acl-long.58} {Increasing
  faithfulness in knowledge-grounded dialogue with controllable features}.
\newblock In \emph{Proceedings of the 59th Annual Meeting of the Association
  for Computational Linguistics and the 11th International Joint Conference on
  Natural Language Processing (Volume 1: Long Papers)}, pages 704--718, Online.
  Association for Computational Linguistics.

\bibitem[{Roller et~al.(2021)Roller, Dinan, Goyal, Ju, Williamson, Liu, Xu,
  Ott, Smith, Boureau, and Weston}]{roller2020recipes}
Stephen Roller, Emily Dinan, Naman Goyal, Da~Ju, Mary Williamson, Yinhan Liu,
  Jing Xu, Myle Ott, Eric~Michael Smith, Y-Lan Boureau, and Jason Weston. 2021.
\newblock \href {https://doi.org/10.18653/v1/2021.eacl-main.24} {Recipes for
  building an open-domain chatbot}.
\newblock In \emph{Proceedings of the 16th Conference of the European Chapter
  of the Association for Computational Linguistics: Main Volume}, pages
  300--325, Online. Association for Computational Linguistics.

\bibitem[{Saha et~al.(2020)Saha, Nie, and Bansal}]{saha2020conjnli}
Swarnadeep Saha, Yixin Nie, and Mohit Bansal. 2020.
\newblock \href {https://doi.org/10.18653/v1/2020.emnlp-main.661} {{C}onj{NLI}:
  Natural language inference over conjunctive sentences}.
\newblock In \emph{Proceedings of the 2020 Conference on Empirical Methods in
  Natural Language Processing (EMNLP)}, pages 8240--8252, Online. Association
  for Computational Linguistics.

\bibitem[{Santhanam et~al.(2021)Santhanam, Hedayatnia, Gella, Padmakumar, Kim,
  Liu, and Hakkani-Tur}]{santhanam2021rome}
Sashank Santhanam, Behnam Hedayatnia, Spandana Gella, Aishwarya Padmakumar,
  Seokhwan Kim, Yang Liu, and Dilek Hakkani-Tur. 2021.
\newblock \href {https://arxiv.org/abs/2110.05456} {Rome was built in 1776: A
  case study on factual correctness in knowledge-grounded response generation}.
\newblock \emph{ArXiv preprint}, abs/2110.05456.

\bibitem[{Shao et~al.(2021)Shao, Geng, Liu, Dai, Yang, Zhe, Bao, and
  Qiu}]{shao2021cpt}
Yunfan Shao, Zhichao Geng, Yitao Liu, Junqi Dai, Fei Yang, Li~Zhe, Hujun Bao,
  and Xipeng Qiu. 2021.
\newblock \href {https://arxiv.org/abs/2109.05729} {Cpt: A pre-trained
  unbalanced transformer for both chinese language understanding and
  generation}.
\newblock \emph{ArXiv preprint}, abs/2109.05729.

\bibitem[{Shuster et~al.(2022)Shuster, Urbanek, Szlam, and
  Weston}]{Shuster2022-is}
Kurt Shuster, Jack Urbanek, Arthur Szlam, and Jason Weston. 2022.
\newblock \href {https://doi.org/10.18653/v1/2022.findings-naacl.182} {Am {I}
  me or you? state-of-the-art dialogue models cannot maintain an identity}.
\newblock In \emph{Findings of the Association for Computational Linguistics:
  NAACL 2022}, pages 2367--2387, Seattle, United States. Association for
  Computational Linguistics.

\bibitem[{Su and Collier(2022)}]{su2022contrastive}
Yixuan Su and Nigel Collier. 2022.
\newblock \href {https://arxiv.org/abs/2210.14140} {Contrastive search is what
  you need for neural text generation}.
\newblock \emph{ArXiv preprint}, abs/2210.14140.

\bibitem[{Wang et~al.(2021)Wang, Li, Zhao, and Yu}]{wang2021naturalconv}
Xiaoyang Wang, Chen Li, Jianqiao Zhao, and Dong Yu. 2021.
\newblock \href {https://ojs.aaai.org/index.php/AAAI/article/view/17649}
  {Naturalconv: {A} chinese dialogue dataset towards multi-turn topic-driven
  conversation}.
\newblock In \emph{Thirty-Fifth {AAAI} Conference on Artificial Intelligence,
  {AAAI} 2021, Thirty-Third Conference on Innovative Applications of Artificial
  Intelligence, {IAAI} 2021, The Eleventh Symposium on Educational Advances in
  Artificial Intelligence, {EAAI} 2021, Virtual Event, February 2-9, 2021},
  pages 14006--14014. {AAAI} Press.

\bibitem[{Wang et~al.(2020)Wang, Ke, Zheng, Huang, Jiang, Zhu, and
  Huang}]{wang2020large}
Yida Wang, Pei Ke, Yinhe Zheng, Kaili Huang, Yong Jiang, Xiaoyan Zhu, and
  Minlie Huang. 2020.
\newblock A large-scale chinese short-text conversation dataset.
\newblock In \emph{Natural Language Processing and Chinese Computing: 9th CCF
  International Conference, NLPCC 2020, Zhengzhou, China, October 14--18, 2020,
  Proceedings, Part I 9}, pages 91--103. Springer.

\bibitem[{Welleck et~al.(2020)Welleck, Kulikov, Roller, Dinan, Cho, and
  Weston}]{welleck2019neural}
Sean Welleck, Ilia Kulikov, Stephen Roller, Emily Dinan, Kyunghyun Cho, and
  Jason Weston. 2020.
\newblock \href {https://openreview.net/forum?id=SJeYe0NtvH} {Neural text
  generation with unlikelihood training}.
\newblock In \emph{8th International Conference on Learning Representations,
  {ICLR} 2020, Addis Ababa, Ethiopia, April 26-30, 2020}. OpenReview.net.

\bibitem[{Welleck et~al.(2019)Welleck, Weston, Szlam, and
  Cho}]{welleck2018dialogue}
Sean Welleck, Jason Weston, Arthur Szlam, and Kyunghyun Cho. 2019.
\newblock \href {https://doi.org/10.18653/v1/P19-1363} {Dialogue natural
  language inference}.
\newblock In \emph{Proceedings of the 57th Annual Meeting of the Association
  for Computational Linguistics}, pages 3731--3741, Florence, Italy.
  Association for Computational Linguistics.

\bibitem[{Wu et~al.(2023)Wu, Wang, Wan, Jiao, and Lyu}]{wu2023chatgpt}
Haoran Wu, Wenxuan Wang, Yuxuan Wan, Wenxiang Jiao, and Michael Lyu. 2023.
\newblock \href {https://arxiv.org/abs/2303.13648} {Chatgpt or grammarly?
  evaluating chatgpt on grammatical error correction benchmark}.
\newblock \emph{ArXiv preprint}, abs/2303.13648.

\bibitem[{Zamani et~al.(2020{\natexlab{a}})Zamani, Dumais, Craswell, Bennett,
  and Lueck}]{Zamani2020-oz}
Hamed Zamani, Susan Dumais, Nick Craswell, Paul Bennett, and Gord Lueck.
  2020{\natexlab{a}}.
\newblock Generating clarifying questions for information retrieval.
\newblock In \emph{Proceedings of The Web Conference 2020}, WWW '20, pages
  418--428, New York, NY, USA. Association for Computing Machinery.

\bibitem[{Zamani et~al.(2020{\natexlab{b}})Zamani, Dumais, Craswell, Bennett,
  and Lueck}]{Zamani2020-nq}
Hamed Zamani, Susan Dumais, Nick Craswell, Paul Bennett, and Gord Lueck.
  2020{\natexlab{b}}.
\newblock Generating clarifying questions for information retrieval.
\newblock In \emph{Proceedings of The Web Conference 2020}, WWW '20, pages
  418--428, New York, NY, USA. Association for Computing Machinery.

\bibitem[{Zhang et~al.(2023)Zhang, Jin, Song, Mi, Chen, and
  Yu}]{zhang2023safeconv}
Mian Zhang, Lifeng Jin, Linfeng Song, Haitao Mi, Wenliang Chen, and Dong Yu.
  2023.
\newblock Safeconv: Explaining and correcting conversational unsafe behavior.
\newblock In \emph{Proceedings of the 61st Annual Meeting of the Association
  for Computational Linguistics (Volume 1: Long Papers)}, pages 22--35.

\bibitem[{Zhang et~al.(2018)Zhang, Dinan, Urbanek, Szlam, Kiela, and
  Weston}]{zhang2018personalizing}
Saizheng Zhang, Emily Dinan, Jack Urbanek, Arthur Szlam, Douwe Kiela, and Jason
  Weston. 2018.
\newblock \href {https://doi.org/10.18653/v1/P18-1205} {Personalizing dialogue
  agents: {I} have a dog, do you have pets too?}
\newblock In \emph{Proceedings of the 56th Annual Meeting of the Association
  for Computational Linguistics (Volume 1: Long Papers)}, pages 2204--2213,
  Melbourne, Australia. Association for Computational Linguistics.

\bibitem[{Zhao et~al.(2019)Zhao, Chen, Zhang, Zhao, Liu, Lu, Chen, Deng, Ju,
  and Du}]{zhao2019uer}
Zhe Zhao, Hui Chen, Jinbin Zhang, Xin Zhao, Tao Liu, Wei Lu, Xi~Chen, Haotang
  Deng, Qi~Ju, and Xiaoyong Du. 2019.
\newblock \href {https://doi.org/10.18653/v1/D19-3041} {{UER}: An open-source
  toolkit for pre-training models}.
\newblock In \emph{Proceedings of the 2019 Conference on Empirical Methods in
  Natural Language Processing and the 9th International Joint Conference on
  Natural Language Processing (EMNLP-IJCNLP): System Demonstrations}, pages
  241--246, Hong Kong, China. Association for Computational Linguistics.

\bibitem[{Zheng et~al.(2022)Zheng, Zhou, Zheng, Peng, Guo, Wu, Niu, Wu, and
  Huang}]{zheng2022cdconv}
Chujie Zheng, Jinfeng Zhou, Yinhe Zheng, Libiao Peng, Zhen Guo, Wenquan Wu,
  Zheng-Yu Niu, Hua Wu, and Minlie Huang. 2022.
\newblock \href {https://aclanthology.org/2022.emnlp-main.2} {{CDC}onv: A
  benchmark for contradiction detection in {C}hinese conversations}.
\newblock In \emph{Proceedings of the 2022 Conference on Empirical Methods in
  Natural Language Processing}, pages 18--29, Abu Dhabi, United Arab Emirates.
  Association for Computational Linguistics.

\end{thebibliography}
